\documentclass{article}
\usepackage{ijcai25}
\usepackage{natbib}
\usepackage{times}
\usepackage{soul}
\usepackage{url}
\usepackage[hidelinks]{hyperref}
\usepackage[utf8]{inputenc}
\usepackage[small]{caption}
\usepackage{graphicx}
\usepackage{amsmath}
\usepackage{makecell}
\usepackage{amsthm}
\usepackage{booktabs}
\usepackage{algorithm}
\usepackage{algorithmic}
\usepackage{comment}
\usepackage[switch]{lineno}
\usepackage{physics}
\usepackage{amssymb}
\usepackage{mathtools}
\usepackage{tikz}
\usetikzlibrary{decorations.pathreplacing}

\usepackage{tablefootnote}
\usepackage{xcolor}

\newcommand{\tb}[1]{\textbf{#1}}
\newcommand{\compactpara}[1]{\noindent\tb{#1}\quad}

\newcommand{\fS}{\mathcal{S}}
\newcommand{\fA}{\mathcal{A}}

\newcommand{\R}{\mathbb{R}}
\newcommand{\E}{\mathbb{E}}

\title{A Survey of In-Context Reinforcement Learning}

\author{
Amir Moeini$^1$\and
Jiuqi Wang$^1$\and
Jacob Beck$^2$\and
Ethan Blaser$^1$\and\\
Shimon Whiteson$^2$\and
Rohan Chandra$^1$\and
Shangtong Zhang$^1$\\
\affiliations
$^1$University of Virginia\\
$^2$University of Oxford\\
\emails
\{amoeini, shangtong\}@virginia.edu
}

\begin{document}

\maketitle

\begin{abstract}
Reinforcement learning (RL) agents typically optimize their policies by performing expensive backward passes to update their network parameters. However, some agents can solve new tasks without updating any parameters by simply conditioning on additional context such as their action-observation histories. This paper surveys work on such behavior, known as in-context reinforcement learning. 
\end{abstract}

\section{Introduction}

Reinforcement learning (RL)~\citep{sutton2018reinforcement} 
is a paradigm for solving sequential decision-making tasks via trial and error.
In RL, an agent incrementally optimizes its policy as it interacts with its environment so as to
maximize a reward signal in the long run.
Here, the policy is a function that maps the agent's observation to the distribution from which its actions are sampled. 
Policies are typically represented with neural networks whose parameters are continually updated during learning.
Selecting actions requires a forward pass through the network.
Updating the parameters usually requires a backward pass, 
which can be expensive for large neural networks in terms of both memory and computation \citep{kingmaAdamMethodStochastic2017}.

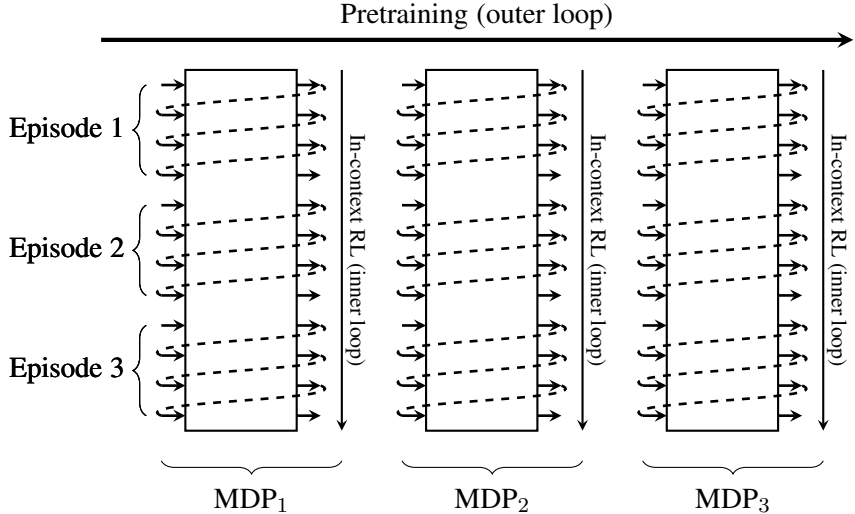
\begin{figure*}[t]
    \centering
    \begin{tikzpicture}[scale=0.4, >=stealth, every node/.style={font=\large}]
        \draw[->, ultra thick] (1,9) -- (26,9)
        node[midway, above] {Pretraining (outer loop)};
    
      \foreach \m in {0,1,2} {
        \pgfmathtruncatemacro{\trial}{\m+1}
        \begin{scope}[xshift={\m*8cm}]
    
        \begin{scope}[rotate around={-90:(8/2,8/2)}, transform shape]
          
          \draw[->, thick] (0,9) -- (12,9)
            node[midway, above] {\Huge In-context RL (inner loop)};
          
          \draw[thick] (0,3.8) rectangle (12,7.5);
          
          \foreach \ep in {0,1,2} {
            \pgfmathsetmacro{\xstart}{4*\ep}      
            \pgfmathtruncatemacro{\episodenumber}{\ep+1}
            
            \foreach \j in {0,1,2,3} {
              \pgfmathsetmacro{\xtoken}{\xstart + 0.5 + \j}
              \draw[->, line width=1pt] (\xtoken,3) -- (\xtoken,3.8);
              \draw[->, line width=1pt] (\xtoken,7.5) -- (\xtoken,8.3);
            }
            
            \foreach \j in {0,1,2} {
              \pgfmathsetmacro{\xout}{\xstart + 0.5 + \j}       
              \pgfmathsetmacro{\xinnext}{\xstart + 0.5 + \j + 1}  
              \draw[dashed, line width=1pt] 
                (\xout,8.3) .. controls (\xout+0.5,9.55) and (\xinnext-0.5,1.75) .. (\xinnext,3);
            }
        }      
    
            \end{scope}
          \draw[decorate,decoration={brace,mirror,amplitude=5pt}]
            (3,-5) -- (9,-5)
            node[midway, yshift=-15pt] {MDP$_\trial$};
    
        \end{scope}

            \draw[decorate,decoration={brace,amplitude=5pt}]
              (2.5,4.5) -- (2.5,7.5)
              node[midway, xshift=-30pt, font=\large] {Episode 1};
            \draw[decorate,decoration={brace,amplitude=5pt}]
              (2.5,.5) -- (2.5,3.5)
              node[midway, xshift=-30pt, font=\large] {Episode 2};
            \draw[decorate,decoration={brace,amplitude=5pt}]
              (2.5,-3.5) -- (2.5,-.5)
              node[midway, xshift=-30pt, font=\large] {Episode 3};
      }
    \end{tikzpicture}
    \caption{Overview of ICRL. After pretraining, the forward pass of the network implements some RL algorithm. The implemented RL algorithm is tested on multiple MDPs. The context in each MDP can span multiple episodes.
    }
    \label{fig: reinforced pretraining}
    \end{figure*}

Some pretrained RL agents, however, can solve new tasks without updating any network parameters.
When evaluating such an agent on a new task,
the input to the agent includes both the current observation and additional context that helps the agent adapt to the new task.
For example, the context may include the agent's history of observations and actions in this new task up to the current time step.
The remarkable ability of such agents to generalize to new tasks using context but without fine-tuning
is hypothesized \citep{duanRL$^2$FastReinforcement2016,laskinContextReinforcementLearning2022} 
to arise from
\tb{the pretrained neural network implementing some (known or unknown) RL algorithm in its forward pass to process the context}.
\tb{We refer to this behavior in the forward pass as in-context reinforcement learning (ICRL)}.
An immediate implication of ICRL is that the agent's performance improves as the task-related information in the context accumulates,
a phenomenon called \emph{in-context improvement}.
Figure~\ref{fig: reinforced pretraining} illustrates ICRL's pipeline.
We follow \citet{sutton2018reinforcement} and define RL algorithms broadly as learning algorithms for solving sequential decision-making problems.
This includes, for example,
imitation learning \citep{abbeelExplorationApprenticeshipLearning2005} and
temporal difference learning (TD, \citet{sutton1988learning}).

We argue that ICRL is important because it enables agents to generalize to new tasks efficiently, requiring only a forward pass without expensive parameter updates. Eliminating the need for parameter updates creates new opportunities to optimize computation and memory requirements for inference \citep{zhu2024survey}.
Furthermore,
there is also evidence that the RL algorithms implemented in the forward pass can potentially be more sample efficient than manually engineered ones \citep{laskinContextReinforcementLearning2022}.

\compactpara{Scope.} 
ICRL falls in the category of black box methods for meta RL \citep{beck2023survey} and dates back to \citet{duanRL$^2$FastReinforcement2016,wangLearningReinforcementLearn2017}.
Early works of ICRL demonstrate only limited out-of-distribution generalization (see Section~\ref{online ctx based methods} for more discussion) and are well surveyed by \citet{beck2023survey}.
We view \citet{laskinContextReinforcementLearning2022} as an important milestone of ICRL since it both coins the term and provides the first demonstration of remarkable out-of-distribution generalization of the pretrained agent.
This survey,
therefore,
focuses on ICRL work after \citet{laskinContextReinforcementLearning2022}.

\compactpara{Taxonomy.}
In this paper,
we survey work on ICRL along different axes.
We start with different pretraining methods, such as supervised pretraining and reinforcement pretraining.
We then examine different methods for constructing context at test time and the demonstrated test time performance.
We then survey recent theoretical advances in understanding ICRL
and, finally, neural network architecture design choices in both empirical and theoretical work on ICRL.

\compactpara{Related Surveys.}
The closest work to this paper is \citet{beck2023survey},
which surveys meta-RL.
\citet{beck2023survey} do not include recent ICRL advances showing strong out-of-distribution generalization since \citet{laskinContextReinforcementLearning2022}.
This survey aims to close this gap and is thus complementary to \citet{beck2023survey}.
See \citet{beck2023survey} for a full treatment of other aspects of meta RL.
The development of ICRL parallels the development of in-context learning (ICL),
where supervised (instead of reinforcement) learning occurs in the forward pass.
See \citet{dongSurveyIncontextLearning2024} for a full treatment of ICL.
Furthermore, some ICRL pretraining methods resemble RL via supervised learning (RvS) in the offline RL community and goal-conditioned RL.
However,
many methods for RvS and goal-conditioned RL, e.g., \citet{chenDecisionTransformerReinforcement2021},
do not demonstrate key properties of ICRL, such as in-context improvement,
and are therefore not included in this survey.
That being said,
we include
RvS works that are able to demonstrate ICRL, e.g., \citet{laskinContextReinforcementLearning2022}.
Nevertheless, see \citet{emmonsRvSWhatEssential2022,wenLargeSequenceModels2023} for a full treatment of RvS, and  \citet{liuGoalConditionedReinforcementLearning2022} for goal-conditioned RL.

\section{Background}
\compactpara{Reinforcement Learning.}
RL typically models environments or tasks as Markov Decision Processes (MDPs). 
An MDP consists of a state space $\fS$, an action space $\fA$, a reward function $r: \fS \times \fA \to \R$, a transition function $p: \fS \times \fS \times \fA \to [0, 1]$, an initial distribution $p_0: \fS \to [0, 1]$, and a discount factor $\gamma \in [0, 1)$.
At time step $0$, 
an initial state $S_0$ is sampled from $p_0$.
At time $t$,
an agent at a state $S_t$ takes an action $A_t$ according to its policy $\pi: \fA \times \fS \to [0, 1]$, i.e., $A_t \sim \pi(\cdot | S_t)$.
The agent then receives a reward $R_{t+1} \doteq r(S_t, A_t)$ and proceeds to a successor state $S_{t+1} \sim p(\cdot | S_t, A_t)$. 
We use $\tau_t$ to denote the trajectory up to time $t$,
defined as, $\tau_t \doteq (S_0, A_0, R_1, S_1, A_1, \dots, S_{t-1}, A_{t-1}, R_t)$.
The value function of the policy $\pi$ is defined as $v_\pi(s) \doteq \E\qty[\sum_{i=1}^\infty \gamma^{i-1} R_{t+i} | S_t = s]$.
There are two fundamental tasks in RL (given an environment).
The first is policy evaluation,
where the goal is to estimate $v_\pi$ for a given $\pi$.
The second is control,
where the goal is to find a policy $\pi$ to maximize the expected total rewards $J(\pi) \doteq \sum_s p_0(s)v_\pi(s)$.
Policy evaluation is the foundation of control.
Given a policy $\pi$ and its value function, 
a better-performing policy $\pi'$ can be computed.
This is called a \emph{policy improvement} step.
Iterating between policy evaluation and policy improvement can generate the optimal policy to maximize $J(\pi)$.

RL methods usually adopt a parameterized policy $\pi_\theta$ (and a parameterized value estimate $v_\theta$).
For example,
$\theta$ could be the parameters of a neural network.
Many RL algorithms have been proposed to update $\theta$ during the agent-environment interaction.
This is referred to as online pretraining.
By contrast, offline pretraining relies on a dataset $D$
consisting of previously collected transition tuples
$D \doteq \qty{(s_i, a_i, r_i, s_i')}_{i=1,\dots,K}$, where $r_i$ is the reward recevied after executing the action $a_i$ in the state $s_i$, and $s_i'$ is the corresponding successor state.
This offline dataset may come from one or more (known or unknown) behavior policies.
The transition tuples may or may not form a complete trajectory. 
Many RL algorithms are proposed to update $\theta$ using $D$ as well.
Notably,
both online and offline pretraining can involve learning a single policy $\pi_\theta$ from one or multiple MDPs.
When a single policy is used for multiple MDPs,
it needs additional input besides the current state to differentiate between and adapt to different MDPs.
After pretraining, whether offline or online,
the policy $\pi_\theta$ is evaluated on the same MDP or one or more new MDPs.

\compactpara{In-Context Reinforcement Learning.}
The key idea of ICRL is to condition the policy both on $S_t$ and some context $C_t$, such that the action $A_t$ is sampled according to $\pi_\theta(\cdot | S_t, C_t)$, rather than solely being conditioned on $S_t$.
This survey considers different ways to construct $C_t$, but a simple example is to use $\tau_t$ as $C_t$. 
Some pretraining methods can generate $\theta$ such that the policy $\pi_\theta(\cdot | S_t, C_t)$ can obtain high rewards in test MDPs that differ from the MDPs seen during pretraining, despite $\theta$ remaining fixed. 
Such generalization is hypothesized to arise because the forward pass of the neural network $\theta$ implements an RL algorithm that learns from the current context $C_t$ \citep{duanRL$^2$FastReinforcement2016,laskinContextReinforcementLearning2022}.
This is called ICRL (for control).
Furthermore, the performance of $\pi_\theta$ improves with the length of the context $C_t$ (provided that $C_t$ always consists of information related to the task), which we call in-context improvement. 
Similarly, RL algorithms for policy evaluation can also be implemented in the forward pass if the neural network for value estimation takes as input both the state and the context \citep{wangTransformersLearnTemporal2024}. 
This is called ICRL for policy evaluation.

\section{Supervised Pretraining} \label{sec: offline ctx based methods}

This section surveys the first class of methods for pretraining the network parameter $\theta$ -- supervised pretraining,
which is commonly done through behavior cloning.
The pretraining objective is typically the log-likelihood $\log \pi_\theta(a^* | s, c)$ or its variants,
where $(s, c)$ is the input state-context pair and $a^*$ is the desired 
output action.
There are multiple ways to construct the input and output.

The most common approach is to concatenate the trajectories from multiple episodes as input (called \emph{cross-episode} input) 
and use the corresponding action at each step as output. 
Various ways are proposed to obtain the episode trajectories for constructing the input.
\citet{laskinContextReinforcementLearning2022} use trajectories generated by some existing RL algorithms across their entire lifecycle. As a result, the input includes trajectories produced by existing RL algorithms during both early pretraining stages (when agents perform poorly) and later stages (when agents achieve strong performance). 
Building on \citet{laskinContextReinforcementLearning2022}, \citet{shiCrossEpisodicCurriculumTransformer2023} propose to fill the context with a curriculum of trajectories.
Specifically, they order the trajectories by task difficulty, 
demonstrator proficiency, 
or episode returns \citep{huangContextDecisionTransformer2024,huangDecisionMambaReinforcement2024,liuEmergentAgenticTransformer2023}. 
Alternative curriculum construction methods include adding decaying noise to expert demonstration trajectories \citep{zismanEmergenceContextReinforcement2023} and using an explicit feature in the trajectory to indicate
whether the current episode is better than the ones before \citep{daiContextExplorationExploitationReinforcement2024}. 
This indicator, called cross-episode return-to-go in \citet{daiContextExplorationExploitationReinforcement2024}, alleviates the need to rank episodes in the context.
All of these methods aim to demonstrate performance improvement in the input-out pairs, encouraging the neural network to implement \tb{some policy improvement algorithm} in the forward pass.
\citet{kirschGeneralPurposeContextLearning2023} show that the performance improvement rate in the input trajectories directly influences how quickly the pretrained agent improves at test time.

Some methods instead emphasize finding trajectories similar to the current task from the offline data and prepending them to the context for better generalization \citep{xuPromptingDecisionTransformer2022,wangHierarchicalPromptDecision2024}. 
These initial trajectories are referred to as a \textit{prompt}. 
This encourages the network to implement \tb{some imitation learning algorithm} in the forward pass to imitate the behavior demonstrated in the context.
This approach is most effective when the prompt is collected by an expert agent \citep{raparthyGeneralizationNewSequential2023,xuPromptingDecisionTransformer2022}.
Consequently, these methods require expert prompts at test time and cannot use their own suboptimal interactions as prompts.
To mitigate this issue, 
\citet{leeSupervisedPretrainingCan2023} construct the prompts using the suboptimal trajectories 
but replace the actions in the trajectories with actions predicted by a performant policy. 

Some works incorporate hindsight information into context to facilitate policy optimization \citep{furutaGeneralizedDecisionTransformer2022}.
Hindsight information is data that can only be inferred from outcomes of future time steps.
A common example is the Return-To-Go (RTG), which represents the sum of rewards from a step until the end of the episode,
\citep{xuPromptingDecisionTransformer2022, wangHierarchicalPromptDecision2024, daiContextExplorationExploitationReinforcement2024}.
However, inspired by how RTG is estimated during testing (Section \ref{sec: test time cond}), some methods replace this with a target RTG, defined as either the maximum RTG in the offline dataset \citep{schmiedLargeRecurrentAction2024} or among episodes in the current context \citep{huangContextDecisionTransformer2024, huangDecisionMambaReinforcement2024}.

To summarize,
while some works do not rely on a curriculum in the context \citep{leeSupervisedPretrainingCan2023}, 
they all rely on multiple trajectories in the context. 
Using a single trajectory in the context \citep{chenDecisionTransformerReinforcement2021} provably prevents an agent from improving on the offline-data-generating policies 
or ``stitching'' suboptimal trajectories to get an optimal one without strong assumptions \citep{brandfonbrenerWhenDoesReturnconditioned2023}.

Sample efficiency during supervised pretraining is also an important research area.
Here we define sample efficiency broadly,
including the amount of data, the amount of expert / optimal policy demonstration,
and the amount of different tasks.
\citet{zismanNGramInductionHeads2024} greatly improve the pretraining sample efficiency of \citet{laskinContextReinforcementLearning2022} by hardcoding $n$-gram induction heads into the transformer and biasing it toward using in-context information \citep{akyurekInContextLanguageLearning2024}. 
\citet{daiContextExplorationExploitationReinforcement2024} use importance sampling to remove the pessimistic bias that keeps the pretrained policy close to the data collection policy, thereby enabling the viable use of suboptimal data collection policies.
Similarly, \citet{dongInContextReinforcementLearning2024} employ a weighted loss, where the weights, based on the observed rewards, act as importance sampling ratios to guide the suboptimal policy toward the optimal policy.
With a different approach, \citet{zismanEmergenceContextReinforcement2023} use a suboptimal policy and add annealing noise to its trajectories to generate learning histories similar to those of \citet{laskinContextReinforcementLearning2022}.
\citet{kirschGeneralPurposeContextLearning2023} construct augmented tasks to improve sample efficiency by randomly projecting the task's observation and action spaces.
While these methods enhance overall sample efficiency, robotic tasks introduce unique challenges. Due to long sequence lengths, full rollouts are inefficient.
\citet{elawadyReLICRecipe64k2024} mitigate this issue by employing partial rollouts that reduce environment interactions.

Finally, there are multiple ways to encode the context before it is presented to the neural network.
For example, in Transformer-based methods, 
the token at each time step can be the stacked embeddings of either  
$(s_t,a_t,r_{t+1},s_{t+1})$ \citep{kirschGeneralPurposeContextLearning2023,leeSupervisedPretrainingCan2023} or $(a_{t-1}, r_t, s_t)$ \citep{laskinContextReinforcementLearning2022,raparthyGeneralizationNewSequential2023,zismanEmergenceContextReinforcement2023}. 
Similarly, action spaces can be embedded in distinct ways. 
For instance, to enable test-time adaptation to varying action spaces, \citet{siniiContextReinforcementLearning2024} project discrete actions into random vector embeddings and 
train the network to output an embedding vector directly.
Then, the action whose embedding is most similar to the network output is executed.

\section{Reinforcement Pretraining} \label{online ctx based methods}

This section surveys the second class of methods for pretraining the network parameter $\theta$ -- reinforcement pretraining. 
Instead of using the log-likelihood loss, reinforcement pretraining uses other established RL algorithms to train the policy $\pi_\theta(a|s,c)$. 
In contrast to supervised pretraining which uses offline datasets, reinforcement pretraining usually involves online environment interactions.

Early ICRL works with reinforcement pretraining include
\citet{duanRL$^2$FastReinforcement2016,wangLearningReinforcementLearn2017,mishraSimpleNeuralAttentive2018,ritterBeenThereDone2018,stadieConsiderationsLearningExplore2019a,zintgrafVariBADVeryGood2020,meloTransformersAreMetaReinforcement2022},
where a sequence model (e.g., an RNN)  parameterizes the policy.
At test time,
the policy takes the agent's online interaction history (usually across multiple episodes) as input and outputs the action,
without any parameter updates.
\emph{This history-dependent policy effectively functions as an RL algorithm, 
as both take the complete history as input and output an action.}
However,
early works in this line of research demonstrate only limited out-of-distribution generalization.
They only demonstrate that the learned history-dependent policy performs well in tasks similar to pretraining tasks. 
One hypothesis for the lack of out-of-distribution generalization in those works is that the pretrained network implements \tb{some task identification algorithm together with certain nearest neighbor matching}. 
In other words,
at test time,
the pretrained network tries to identify pretraining tasks that are similar to the test task (based on the entire online interaction history in the test task) and acts as if the test task was one of those similar pretraining tasks.
Those task identification works are well surveyed by \citet{beck2023survey}.

In this paper, 
we instead focus on more recent advances in ICRL, where pretrained networks demonstrate stronger out-of-distribution generalization by implementing more advanced RL algorithms in the forward pass.
These include \citet{grigsbyAMAGO2BreakingMultiTask2024,grigsbyAMAGOScalableContext2024,luStructuredStateSpace2023,teamHumanTimescaleAdaptationOpenEnded2023,parkLLMAgentsHave2024,wangTransformersLearnTemporal2024,elawadyReLICRecipe64k2024,xuMetaReinforcementLearningRobust2024,cookArtificialGenerationalIntelligence2024}.
Although their pretraining is still performed by (modification of) standard RL algorithms (Table~\ref{fig: reinforced pretraining}),
using long-context neural networks such as Transformers to parameterize the policy introduces substantial learning stability challenges \citep{grigsbyAMAGOScalableContext2024}.
To improve stability, several modifications to the pretraining process have been proposed. 
\citet{grigsbyAMAGOScalableContext2024} employ multiple discount rates simultaneously to stabilize long-horizon credit assignment. \citet{teamHumanTimescaleAdaptationOpenEnded2023} introduce a task selection strategy that prioritizes tasks slightly beyond the agent's current expertise, significantly enhancing sample efficiency. For scenarios involving learning across tasks with highly varied return scales, \citet{grigsbyAMAGO2BreakingMultiTask2024} utilize actor-critic objectives decoupled from return magnitudes, thereby improving convergence.
That being said,
why the recent works are able to demonstrate stronger generalization remains an open problem (Section~\ref{sec open problem}).

\begin{table}[h]
    \centering
    \begin{tabular}{|l|l|}
    \hline
    \textbf{Method} & \textbf{Pretraining Algorithm} \\ \hline
    \citet{grigsbyAMAGOScalableContext2024} & \makecell[l]{Modified DDPG \\ \citep{lillicrap2019continuouscontroldeepreinforcement}} \\ \hline
    \citet{grigsbyAMAGO2BreakingMultiTask2024} & Modified DDPG \\ \hline
    \citet{luStructuredStateSpace2023} & Muesli \citep{hessel2021muesli}\\ \hline
    \citet{teamHumanTimescaleAdaptationOpenEnded2023} & Muesli \\ \hline
    \citet{elawadyReLICRecipe64k2024} & \makecell[l]{Modified PPO \\ \citep{schulman2017proximalpolicyoptimizationalgorithms} }\\ \hline
    \citet{wangTransformersLearnTemporal2024} & TD \\ \hline
    \citet{parkLLMAgentsHave2024} & Regret minimization \\ \hline
    \citet{cookArtificialGenerationalIntelligence2024} & PPO \\\hline
    \citet{xuMetaReinforcementLearningRobust2024} & \makecell[l]{Modified DQN\\ \citep{mnih2015human}} \\\hline
    \end{tabular}
    \caption{Algorithms used for reinforcement pretraining.}
    \label{tab: rl_pretraining_outerloop}
\end{table}

\section{Test Time Context} \label{sec: test time cond}
Having surveyed the two main pretraining paradigms,
we now turn to test time design choices, beginning with the context construction.
While context construction is often similar during both pretraining and testing,
some information provided in the context construction during pretraining is not available at test time.
This section starts with examining how to address these differences.

One example of such information is expert demonstrations. 
Methods using such demonstrations in pretraining \citep{rakellyEfficientOffPolicyMetaReinforcement2019,leeSupervisedPretrainingCan2023,wangHierarchicalPromptDecision2024} often experience significant performance drops when prompted with suboptimal interactions in test time.
This decline occurs because the model has limited exposure to the near-optimal trajectory space during testing, leading to a mismatch between the context distribution in pretraining and testing.

If at test time the agent outputs actions using only offline trajectories as context,
the performance will heavily depend on the quality of the offline demonstrations \citep{leeSupervisedPretrainingCan2023}.
This is also the case if the agent is expected to output good actions after obtaining only one or a few online trajectories \citep{raparthyGeneralizationNewSequential2023,wangHierarchicalPromptDecision2024}.
But when the agent is allowed to interact more extensively with the environment at test time before it is expected to output good actions, 
the need for expert demonstrations in the context is reduced \citep{laskinContextReinforcementLearning2022,huangContextDecisionTransformer2024,huangDecisionMambaReinforcement2024,siniiContextReinforcementLearning2024,kirschGeneralPurposeContextLearning2023,leeSupervisedPretrainingCan2023}. 
\citet{leeSupervisedPretrainingCan2023} demonstrate the trade-off between the allowed online interaction budget and the need for expert demonstrations. 

Another example of context information that is not available at test time is RTG from Section \ref{sec: offline ctx based methods}. 
Various methods are used to estimate RTG at test time.
\citet{huangDecisionMambaReinforcement2024,huangContextDecisionTransformer2024,xuPromptingDecisionTransformer2022,schmiedLargeRecurrentAction2024,wangHierarchicalPromptDecision2024}
approximate RTG once per task based on the offline trajectories.
At the beginning of the episode,
an initial RTG is given.
This RTG is iteratively updated based on the observed reward.
Alternatively, \citet{daiContextExplorationExploitationReinforcement2024} train a secondary network to predict RTG dynamically based on the interaction history. 

During testing, methods that do not rely on demonstrations must learn the task solely from their own previous interactions. However, we can selectively choose what to include in the context. For instance, \citet{cookArtificialGenerationalIntelligence2024} divide the total interaction horizon into generations, with each generation comprising several agents. These agents systematically use the best-performing agent from the previous generation to generate interactions that are then incorporated into the current context. This approach allows the current agent to build upon prior experience. The framework, referred to as cultural accumulation, achieves superior test-time performance scaling compared to the base single-generation method.

\section{Test Time Performance} \label{sec: generalization}
In this section, 
we survey the test-time performance of pretrained agents from two aspects, generalization and sample efficiency.
Since comparing generalization across different benchmarks is challenging, we consider generalization benchmark by benchmark.
Notably, unlike goal-conditioned methods that explicitly condition the pretrained agent on the test task type, 
the ICRL agent must infer it implicitly by itself. 

The first remarkable out-of-distribution generalization in ICRL 
is demonstrated by \citet{laskinContextReinforcementLearning2022} in multi-armed bandit problems.
Their pretrained agents learn new bandit problems with adversarial rewards (i.e., engineered so that pretraining-optimal policies perform poorly) and achieve regret nearly equivalent to standard bandit algorithms that involve parameter updates.

ICRL’s out-of-distribution generalization improves as models, pretraining duration, and experience diversity scale \citep{teamHumanTimescaleAdaptationOpenEnded2023,kirschGeneralPurposeContextLearning2023}. 
For instance,
\citet{teamHumanTimescaleAdaptationOpenEnded2023} propose XLand 2.0, a procedurally generated 3D environment featuring diverse goals, rules, and configurations. They demonstrate generalization on this challenging benchmark using ICRL, enabled by large-scale supervised pretraining with a curriculum and other improvements. 

Other commonly used benchmarks in the ICRL literature include
Dark Room \citep{laskinContextReinforcementLearning2022}, DMLab Watermaze \citep{laskinContextReinforcementLearning2022}, Procgen \citep{cobbe2019procgen}, Meta-World \citep{yuMetaWorldBenchmarkEvaluation2021}, and Mujoco Control \citep{todorov2012mujoco}. 
Each benchmark requires a specific form of generalization which is demonstrated by different works.
Test tasks can differ from pretraining tasks across various task-dependent factors (such as goal location or object types). Accordingly, the difficulty of generalization can be better understood by considering the types of factors and the extent of variations \citep{kirkSurveyZeroshotGeneralisation2023}.

To succeed in the Dark Room benchmark and its variants, the agent should learn to efficiently find held-out (i.e., not used during pretraining) invisible goal locations.
Each unique goal location (or combination of them in the key-to-door variant) represents a new task. 
This generalization is demonstrated by \citet{laskinContextReinforcementLearning2022,leeSupervisedPretrainingCan2023,huangContextDecisionTransformer2024,huangDecisionMambaReinforcement2024,siniiContextReinforcementLearning2024,zismanEmergenceContextReinforcement2023,zismanNGramInductionHeads2024,kirschGeneralPurposeContextLearning2023,daiContextExplorationExploitationReinforcement2024,grigsbyAMAGOScalableContext2024,elawadyReLICRecipe64k2024} to different degrees. Notably, \citet{grigsbyAMAGOScalableContext2024} adapts to a new key-to-door environment in only 300 interactions.

Similarly, to succeed in DMLab Watermaze, where the input to the agent is raw pixels, 
the agent needs to find a trapdoor in new locations of a maze.
This generalization is demonstrated by
\citet{laskinContextReinforcementLearning2022,zismanEmergenceContextReinforcement2023,shiCrossEpisodicCurriculumTransformer2023,ritterBeenThereDone2018} to different degrees.

To succeed in the Mujoco Control benchmark,
the agent must control simulated robots to achieve given tasks (e.g., make a HalfCheetah run or an Ant navigate)
with variations in 
dynamics (e.g., altered friction or mass) or target parameters (e.g., desired speed or direction).
This generalization is demonstrated by
\citet{xuPromptingDecisionTransformer2022,wangHierarchicalPromptDecision2024,grigsbyAMAGOScalableContext2024,mishraSimpleNeuralAttentive2018,meloTransformersAreMetaReinforcement2022,kirschGeneralPurposeContextLearning2023} to different degrees.
In particular,
\citet{kirschGeneralPurposeContextLearning2023} show that after pretraining on Ant tasks,
the agent can solve the Cartpole task in DeepMind Control Suite.

Procgen is a benchmark consisting of 16 procedurally generated 2D games (e.g., platformers, puzzles) with pixel observations. To demonstrate generalization, the agent should learn held-out games \citep{raparthyGeneralizationNewSequential2023}. The games differ across many factors (such as objects, objectives, and types of movement), which have proved difficult to generalize to,
especially when expert demonstrations are not available during the test.
\citet{grigsbyAMAGO2BreakingMultiTask2024,schmiedLargeRecurrentAction2024} show initial progress in an easier setting,  
where they test 
on the same pretraining games with limited modifications (e.g., changes to the starting location or textures)

Meta-World consists of robotic manipulation challenges. 
In Meta-Learning 1 (ML1), the variations are continuous (e.g., different object or goal positions) within a single manipulation category. 
By contrast, ML45 uses 45 manipulation categories (e.g., opening drawers or turning faucets) for pretraining and 5 new categories for testing. Several studies have shown models generalizing on ML1 \citep{xuPromptingDecisionTransformer2022,wangHierarchicalPromptDecision2024,grigsbyAMAGOScalableContext2024,mishraSimpleNeuralAttentive2018,meloTransformersAreMetaReinforcement2022}, and \citet{grigsbyAMAGO2BreakingMultiTask2024} show generalization on the 45 pretraining manipulation categories of ML45 in a setting similar to the limited one described earlier for Procgen.

We now turn to the sample efficiency of the pretrained agents in the test time.
\citet{laskinContextReinforcementLearning2022} demonstrate that 
a pretrained network with fixed parameters needs fewer samples at test time to achieve similar performance to that of baseline RL algorithms that require gradient updates.
\citet{kirschGeneralPurposeContextLearning2023} successfully control test-time sample efficiency by manipulating how much an episode improves upon the previous one when constructing the cross-episode pretraining contexts. 
\citet{leeSupervisedPretrainingCan2023} show that the forward pass of their pretrained network is an efficient implementation of posterior sampling, a sample-efficient RL algorithm, under specific conditions during the test. 
Likewise, \citet{xuMetaReinforcementLearningRobust2024} propose an end-to-end framework for learning an agent that performs Bayesian inference in context, thereby improving test-time sample efficiency on out-of-distribution tasks.
Sample efficiency remains a challenge in sparse-reward tasks when the agent is not sufficiently biased toward thorough exploration. \citet{stadieConsiderationsLearningExplore2019a,normanFirstExploreThenExploit2024} propose addressing this issue by modifying the objective to maximize only the cumulative reward of later exploitive episodes, 
thereby allowing the initial explorative episodes to focus on better exploration for subsequent exploitive episodes.

\section{Theory}\label{sec: theory}
We now consider recent advances in the theoretical understanding of ICRL.

\compactpara{Supervised Pretraining.}
Supervised pretraining can be understood through the lens of behavior cloning. 
In canonical behavior cloning,
the goal is to learn a policy.
The policy usually depends on only the current state or the history within the current episode.
In supervised pretraining, the goal is to learn an algorithm similar to the source algorithm used to generate the offline dataset. 
In other words, to learn a policy that depends on the entire history of previous episodes.
\citet{linTransformersDecisionMakers2024} derive a general bound on the behavioral similarity and performance gap between the learned algorithm (in the forward pass of the neural network) and the source algorithm.
Behavioral similarity is the similarity between the action distributions generated by the learned and source algorithms given the same input.
The performance gap is their difference in episode return.
They further demonstrate how Transformers can approximate LinUCB~\citep{chu2011contextual}, Thompson sampling~\citep{russo2018tutorial}, and UCB-VI~\citep{azar2017minimax} in the forward pass and provide the respective regret bounds.
A follow-up work by~\citet{shiTransformersGamePlayers2024} presents an analogous behavioral similarity guarantee of supervised pretraining for decentralized and centralized learning in two-player zero-sum Markov games.
\citet{shiTransformersGamePlayers2024} further prove by construction that there exist Transformers that can realize V-learning~\citep{jin2024vlean} for decentralized learning and VI-ULCB~\citep{bai2020vi-ulcb} for centralized learning in the forward passes,
accompanied with upper bounds of the approximation error of Nash equilibria for both settings.

\compactpara{Reinforcement Pretraining.}
\citet{parkLLMAgentsHave2024} propose to pretrain language models directly by minimizing regret without requiring action labels.
They show theoretically that 
by minimizing regret with sufficiently many pretraining trajectories,
the pretrained language models can demonstrate no-regret learning at test time.
Lastly, they prove that the global minimizer of the (surrogate) regret loss with a single-layer linear attention transformer implements the known no-regret algorithm Follow-The-Regularized-Leader (FTRL)~\citep{shalev2007online} in the forward pass.
\citet{wangTransformersLearnTemporal2024} consider ICRL for policy evaluation.
They prove by construction that Transformers can precisely implement temporal difference methods in the forward pass for policy evaluation, including TD($\lambda$) \citep{sutton1988learning} and average reward TD \citep{tsitsiklis1999average}.
They also show that those parameters naturally emerge when they train a value estimation transformer with TD on multiple policy evaluation tasks.
Theoretical understanding of this emergence of TD is provided from an invariant set perspective.

\section{Architectures}
\label{sec arc}
A central design choice in ICRL is the architecture of the neural network used to process context. 
The neural network must be able to handle long context lengths,
often containing multiple episodes of interaction, 
and effectively use information from many past interactions.

Although earlier meta RL works \citep{duanRL$^2$FastReinforcement2016,wangLearningReinforcementLearn2017,ritterBeenThereDone2018} use RNN and its variants to parameterize history-dependent policies, 
most surveyed ICRL works employ a causal transformer backbone \citep{laskinContextReinforcementLearning2022, leeSupervisedPretrainingCan2023, raparthyGeneralizationNewSequential2023, siniiContextReinforcementLearning2024, zismanEmergenceContextReinforcement2023, shiCrossEpisodicCurriculumTransformer2023, kirschGeneralPurposeContextLearning2023, xuPromptingDecisionTransformer2022, grigsbyAMAGO2BreakingMultiTask2024, grigsbyAMAGOScalableContext2024, teamHumanTimescaleAdaptationOpenEnded2023, meloTransformersAreMetaReinforcement2022, elawadyReLICRecipe64k2024, zismanNGramInductionHeads2024},
given transformer's demonstrated efficacy in handling long sequences \citep{vaswaniAttentionAllYou2023}. 
However, the inference time of Transformers is quadratic w.r.t.\ the input length.
To speed it up,
state space models \citep{guMambaLinearTimeSequence2024}, whose inference time is linear, are used \citep{cookArtificialGenerationalIntelligence2024}.
\citet{huangDecisionMambaReinforcement2024} employ a state space model for their high-level decision maker, which processes long histories, and a transformer for their low-level decision maker, which processes shorter sequences.
\citet{luStructuredStateSpace2023} modify an existing state space model, S5 \citep{smith2023simplified}, such that it becomes compatible with cross-episode context.
\citet{schmiedLargeRecurrentAction2024} use an xLSTM \citep{beckXLSTMExtendedLong2024} for similar purposes.

Hierarchical structures are also designed with different objectives. For instance, 
\citet{wangHierarchicalPromptDecision2024} improve \citet{xuPromptingDecisionTransformer2022} by incorporating additional modules that extract both task-level and step-specific 
prompts relevant to the current task and step, which are then used to augment the context provided to a Transformer. 
\citet{daiContextExplorationExploitationReinforcement2024} use a secondary network to predict the RTG required for the context during inference.
To improve computational efficiency by processing fewer tokens in a transformer without sacrificing overall historical information, \citet{huangContextDecisionTransformer2024,huangDecisionMambaReinforcement2024} split decision-making into two levels. Specifically, the high-level module processes tokens sampled at fixed intervals, while the low-level module predicts the intervening tokens corresponding to each high-level token.

Compared to supervised pretraining, 
reinforcement pretraining introduces engineering challenges regarding stability during pretraining \citep{grigsbyAMAGOScalableContext2024}. 
To address these challenges, \citet{grigsbyAMAGOScalableContext2024} share a single sequence model across both actor and critic networks and demonstrate that preventing the critic's objective from minimizing the actor's objective can ensure pretraining stability.
They also modify 
the transformer architecture to preserve plasticity over long pretraining durations and avoid performance collapse, also adopted by \citet{xuMetaReinforcementLearningRobust2024}.
Similarly, \citet{elawadyReLICRecipe64k2024} append learnable key and value vectors as ``sinks'' to the transformer's attention mechanism to provide the flexibility of not attending to any input token. This modification makes learning faster and more stable in scenarios involving long but low-information observation sequences, such as those encountered in robotics \citep{elawadyReLICRecipe64k2024}.

Regarding theoretical analysis, having a full-sized multi-layer transformer with arbitrary nonlinear activations is prohibitively challenging due to the complexity of the network structure. 
\citet{linTransformersDecisionMakers2024} and~\citet{shiTransformersGamePlayers2024} use masked attentions with ReLU activations,
and \citet{parkLLMAgentsHave2024,wangTransformersLearnTemporal2024} use linear attentions.

\section{Open Problems and Opportunities}
\label{sec open problem}

ICRL is an emerging area with many open problems.
First, we draw attention to multi-agent RL. 
Generalization to unseen agents (teammates or opponents) during the deployment time is a fundamental challenge in multi-agent RL and meta RL has been applied to address this challenge \citep{charakorn2021learning,gerstgrasser2022meta}.
However,
the demonstrated generalization is only limited in small-scale problems and only limited out-of-distribution generalization is demonstrated.
Recent advances in multi-agent RL with large sequence models \citep{meng2023offline} provide a new opportunity to address this challenge with ICRL.

Second, 
we draw attention to robotics.
ICRL is now only demonstrated in simulated environments.
The sim-to-real gap is a well-known generalization challenge in robotics.
It is a promising direction to investigate whether ICRL will emerge in recent internet-scale robot pretraining \citep{rt2} 
and whether ICRL can help close the sim-to-real gap.

Lastly, we draw attention to reinforcement pretraining. 
\citet{krishnamurthyCanLargeLanguage2024} criticize ICRL saying ``they are explicitly trained to produce this behavior using data from reinforcement learning agents or expert demonstrations on related tasks.'' 
This criticism might be true for supervised pretraining but clearly does not hold for reinforcement pretraining,
where the network is only trained to output good actions or to output good value estimates without constraints on how the network achieves this.
The network itself discovers that implementing certain RL algorithms in the forward pass is a good solution.
In this sense,
ICRL truly emerges during reinforcement pretraining.
Existing theoretical analyses on reinforcement pretraining \citep{parkLLMAgentsHave2024,wangTransformersLearnTemporal2024} use simplified models and simplified pretraining algorithms.
Fully white-boxing the emergence of ICRL during reinforcement pretraining in more realistic settings
remains an open problem,
both theoretically and empirically. 

\section{Conclusion}
This paper presented the first comprehensive survey of ICRL,
an emerging and flourishing area.
We surveyed ICRL from different aspects,
including both pretraining and testing,
both empirical and theoretical analyses.
We hope this survey will stimulate the growth of the ICRL community.

\section*{Acknowledgements}
This work is supported in part by the US National Science Foundation (NSF) under grants III-2128019 and SLES-2331904. 
EB acknowledges support from the NSF Graduate Research Fellowship (NSF-GRFP) under award 1842490. 
This work is also supported in part by the Coastal Virginia Center for Cyber Innovation (COVA CCI)
and the Commonwealth Cyber Initiative (CCI), an investment in the advancement of cyber research and development, innovation, and workforce development. For more information about CCI, visit \url{www.covacci.org} and \url{www.cyberinitiative.org}.

\bibliographystyle{named}
\bibliography{bibliography}

\end{document}